\documentclass[final]{igarss}

\usepackage{times}
\usepackage{epsfig}
\usepackage{amssymb}
\usepackage{algorithm}
\usepackage{amsmath}
\usepackage{amssymb}
\usepackage{bbding} 
\usepackage{booktabs}
\usepackage{multirow}
\usepackage{algorithm}
\usepackage[noend]{algpseudocode}

\usepackage{bm}
\usepackage{bbm}
\usepackage{cite}
\usepackage{graphicx}
\usepackage[normalem]{ulem}
\useunder{\uline}{\ul}{}
\usepackage{subfig}
\newcommand{\tl}[1]{\title{\vspace{-40pt}\MakeUppercase{#1}\vspace{-10pt}}}

\usepackage[pagebackref,breaklinks,colorlinks]{hyperref}
\usepackage[a4paper, left=1.8cm, right=1.8cm, top=2cm, bottom=2cm]{geometry}

\graphicspath{{./figs/}{./figs/result/}}
\providecommand{\keywords}[1]
{{
\small
\textbf{\textit{Index Terms}}--- #1
}}

\begin{document}

\tl{TDiffDe: A Truncated Diffusion Model for Remote Sensing Hyperspectral Image Denoising}

\author{\textit{Jiang He}$^1$, \textit{Graduate Student Member},~IEEE, \textit{Yajie Li}$^1$, \textit{Jie Li}$^1$, \textit{Member},~IEEE, \\\textit{Qiangqiang Yuan}$^1$, \textit{Member},~IEEE\\ \\
$^1$School of Geodesy and Geomatics, Wuhan University, Wuhan, China}

\maketitle

\begin{abstract}
   Hyperspectral images play a crucial role in precision agriculture, environmental monitoring or ecological analysis. However, due to sensor equipment and the imaging environment, the observed hyperspectral images are often inevitably corrupted by various noise. In this study, we proposed a truncated diffusion model, called TDiffDe, to recover the useful information in hyperspectral images gradually. Rather than starting from a pure noise, the input data contains image information in hyperspectral image denoising. Thus, we cut the trained diffusion model from small steps to avoid the destroy of valid information. Results on two datasets with additive white Gaussian noise and hybrid noise all show the reliability and superiority of TDiffDe. Moreover, TDiffDe only require Gaussian noise as guide rather than specific degradations for multi–type noise during training process.
\end{abstract}


\section{Introduction}
%
%
%
%
Hyperspectral (HS) images, due to their rich spectral information, have been widely applied in dealing with various problems, such as classification, object detection, tracking, and damage detection. However, due to the lack of imaging energy caused by the narrow band of the imaging spectrometer, the observed HS images are often inevitably corrupted by serious noise, stripes, and deadlines, which degrades the HS images and limit their applications greatly. 

As the modeling to image structure attracts growing concern, total variation (TV) and low–rank (LR) regularization are widely used to exploit the HS priors. 
TV regularization can effectively preserve the edges and enhance the spatial details. For example, Yuan \etal \cite{ssahtv} proposed a spectral–spatial adaptive total variation (SSAHTV).
As for LR–based models, Zhang \etal \cite{lrmr} proposed low–rank matrix recovery (LRMR) model. 
The LR–based methods can effectively remove sparse noise, while the TV–based methods can properly preserve the high–frequency information from strong Gaussian noise. Combining LR–based with TV–based methods can further improve the ability of the model to represent HS noise. He \etal \cite{lrtv} proposed the total variation regularized low–rank matrix factorization (LRTV) model. Chen \etal \cite{lrtdtv} proposed a factor group sparsity–regularized low–rank approximation and calculated group sparsity on the subspace bases. 

The spatial–spectral information from multiple HS priors provides a more accurate representation of noise for the model. However, their parameters should be adjusted with the human intervention. Moreover, there are a large amount of prior regularization representing different HS properties. Deep learning is a good solution to learn the HS priors from training data implicitly. 

Yuan \etal \cite{hsid} early designed a spatial–spectral deep residual convolutional neural network to explore the high correlation between the adjacent bands. 
Considering the self–similarity and long–range interaction, Fu \etal \cite{nssnn} proposed a non–local self–similarity neural network (NSSNN).
As 3D convolutions can directly address HS images in the spatial–spectral domain, Liu \etal \cite{3dadcnn} proposed a spatial–spectral deep CNNs with 3D atrous convolution. 
Moreover, Wei \etal \cite{qrnn3d} proposed a 3D quasi–Recurrent Neural Network (QRNN3D) and has achieved impressive denoising performance. 

Although the existing deep learning–based denoising methods have achieved ideal performance, they directly remove noise at one step, which is efficient but rude. Diffusion model \cite{ddpm} is a increasingly popular latent variable model. In addition, the step–by–step restoration allows models to recover useful information gradually. In this paper, we proposed a truncated diffusion model (TDiffDe) to explore the full potential of diffusion model in HS image denoising.
\begin{figure*}
	\centering
	\includegraphics[height=140pt]{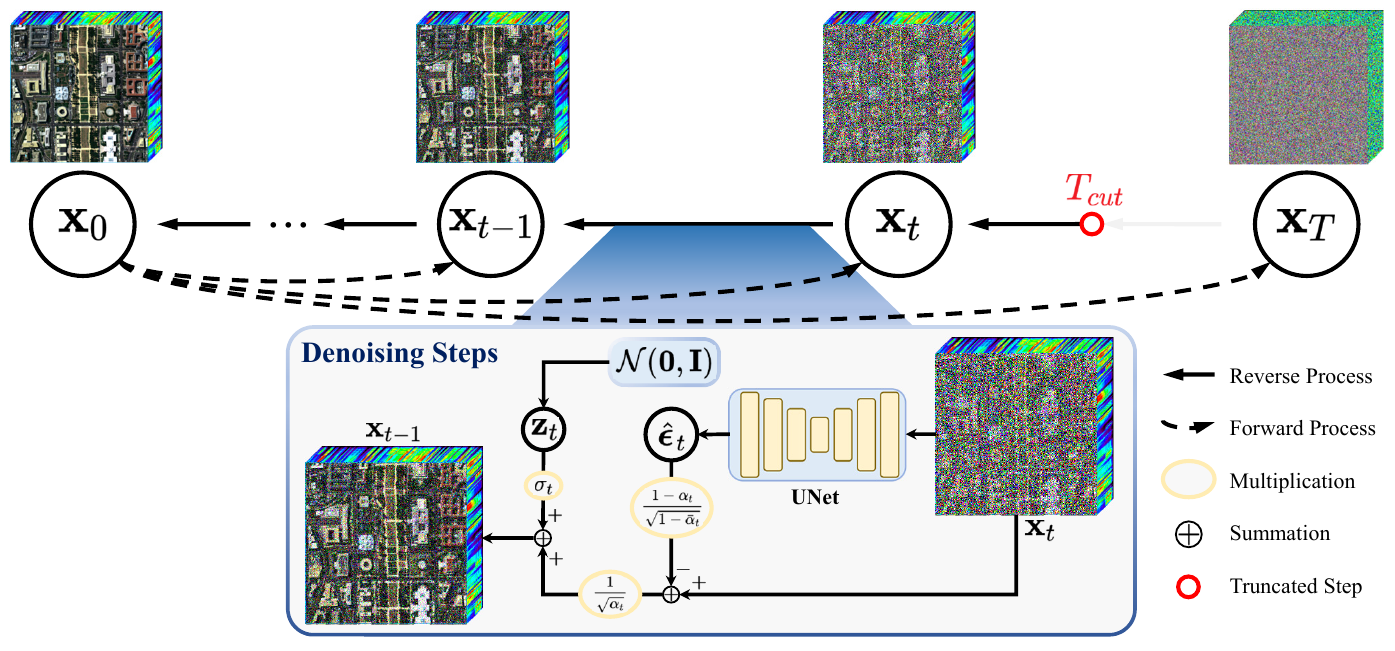}
	\caption{The framework of the proposed TDiffDe.}\label{framework}
\end{figure*}
\begin{itemize} \item As far as we know, this paper is an early work that introduce diffusion model into remote sensing hyperspectral image denoising. \item Not directly employing the reverse processing, this paper truncates the denoising process in diffusion model to ensure that the estimate noise is at the real hyperspectral noise level. \item The proposed truncated diffusion model does not require the real hybrid noise or simulated hybrid noise, the model can learn the noise mining capability adaptively during the forward diffusion training. 
\end{itemize}

The remaining part of the paper is organized as follows. Section 2 describes the diffusion model and the ditails of TDiffDe. Section 3 shows the experiments on some data sets and presents some discussions. Finally, conclusions are given in Section 4. 

\section{The proposed TDiffDe}

Diffusion model is famous for its strong image generation capability and consists of forward process and reverse process. In this work, we proposed a truncated diffusion model named TDiffDe to address hyperspectral image denoising problem, as shown in Fig.~\ref{framework}. 

\subsection{Forward diffusion training}
Given a data sampled from a real data distribution $\mathbf{x}_0 \sim q(\mathbf{x})$, the forward diffusion process can be defined as adding Gaussian noise to the $\mathbf{x}_0$ in $T$ steps gradually and producing a noisy set $\mathbf{x}_1, \dots, \mathbf{x}_T$. The intensity of the added noise is controlled by $\{\beta_t \in (0, 1)\}_{t=1}^T$:
\begin{equation}
\begin{aligned}
q(\mathbf{x}_t \vert \mathbf{x}_{t–1}) &= \mathcal{N}(\mathbf{x}_t; \sqrt{1 – \beta_t} \mathbf{x}_{t–1}, \beta_t\mathbf{I}) \\
q(\mathbf{x}_{1:T} \vert \mathbf{x}_0) &= \prod^T_{t=1} q(\mathbf{x}_t \vert \mathbf{x}_{t–1})
\end{aligned}
\label{eq1}
\end{equation}
where $\mathcal{N}$ denotes the normal distribution and $\mathbf{I}$ means the unit matrix. Eq \ref{eq1} shows the relationship between $\mathbf{x}_{t}$ and $\mathbf{x}_{t–1}$. To directly obtain $\mathbf{x}_{t}$ at any step $t$, we reparameterize $\alpha_t = 1 – \beta_t$ and $\bar{\alpha}_t = \prod_{i=1}^t \alpha_i$. With the additivity principle of Gaussian distributions, $q(\mathbf{x}_t \vert \mathbf{x}_0)$ can be formulated as:
\begin{equation}
q(\mathbf{x}_t \vert \mathbf{x}_0) = \mathcal{N}(\mathbf{x}_t; \sqrt{\bar{\alpha}_t} \mathbf{x}_0, (1 – \bar{\alpha}_t)\mathbf{I})
\label{eq2}
\end{equation}

In the real test, to accurately estimate $\boldsymbol{\epsilon}_t$ in a implicit manner, a deep learning–based model $\boldsymbol{\epsilon}_\theta$ is trained to predict $\boldsymbol{\epsilon}_t$ in the forward diffusion process. The training algorithm is shown in Algorithm \ref{al1}. In this paper, we utilized U–Net as the noise predictor \cite{ddpm}.
\begin{algorithm}[tb]
	\caption{Training of TDiffDe}
	\label{al1}
	\begin{algorithmic}[1]
		\Require
		{Hyperspectral image training set $\mathbf{X}_0$, the diffusion step $T$, the hyperparameters $\{\beta_1,\dots,\beta_T\}$}
		\Ensure {The CNN–based noise predictor $\boldsymbol{\epsilon}_\theta$}
		\State{Calculate $\{\alpha_1,\dots,\alpha_T\}$ following $\alpha_t = 1 – \beta_t$}
		\Repeat
		\State{Randomly select a hyperspectral image $\mathbf{x}_0$ from $\mathbf{X}_0$}
		\State{Randomly select a diffusion step $t$ from $\{1,\dots,T\}$}
		\State{Randomly sample a Gaussian noise $\boldsymbol{\epsilon}_t \sim \mathcal{N}(\mathbf{0}, \mathbf{I})$}
		\State{Calculate $\bar{\alpha}_t = \prod_{i=1}^t \alpha_i$}
		\State{Generate $\mathbf{x}_{in}=\sqrt{\bar{\alpha}_t}\mathbf{x}_0 + \sqrt{1 – \bar{\alpha}_t}\boldsymbol{\epsilon}_t$}
		\State{Feed $\mathbf{x}_{in},~t$ into $\boldsymbol{\epsilon}_\theta$ and obtain the estimated noise $\hat{\boldsymbol{\epsilon}}_t$}
		\State{Compute the $L_2$ loss between $\boldsymbol{\epsilon}_t$ and $\hat{\boldsymbol{\epsilon}}_t$}
		\State{Optimize the CNN–based noise predictor $\boldsymbol{\epsilon}_\theta$}
		\Until{convergence}
	\end{algorithmic}
\end{algorithm}
\subsection{Reverse denoising processing}
Starting at a Gaussian noise input $\mathbf{x}_T \sim \mathcal{N}(\mathbf{0}, \mathbf{I})$, the reverse process is to generate images in the target data distribution $q(\mathbf{x}_0)$. The reverse denoising process can be defined as a Markov chain with learned Gaussian transitions:
\begin{equation}
\begin{aligned}
p_\theta(\mathbf{x}_{0:T}) &= p(\mathbf{x}_T) \prod^T_{t=1} p_\theta(\mathbf{x}_{t–1} \vert \mathbf{x}_t) \\	
p_\theta(\mathbf{x}_{t–1} \vert \mathbf{x}_t) &= \mathcal{N}(\mathbf{x}_{t–1}; \boldsymbol{\mu}_\theta(\mathbf{x}_t, t), \boldsymbol{\Sigma}_\theta(\mathbf{x}_t, t))
\end{aligned}
\end{equation}
where $\boldsymbol{\mu}_\theta(\mathbf{x}_t, t)$ and $\boldsymbol{\Sigma}_\theta(\mathbf{x}_t, t)$ is the mean and variance. When conditioned on $\mathbf{x}_0$, based on \emph{Bayes Rule}, the reverse conditional probability $p_\theta(\mathbf{x}_{t–1} \vert \mathbf{x}_t)$ can be formulated as:
\begin{equation}
q(\mathbf{x}_{t–1} \vert \mathbf{x}_t, \mathbf{x}_0) = \mathcal{N}(\mathbf{x}_{t–1}; \tilde{\boldsymbol{\mu}}(\mathbf{x}_t, \mathbf{x}_0), \tilde{\beta}_t \mathbf{I})
\end{equation}
where $\tilde{\beta}_t$ and $\tilde{\boldsymbol{\mu}}_t (\mathbf{x}_t, \mathbf{x}_0)$ are parameterized as follows:
\begin{equation}
\begin{aligned}
\tilde{\boldsymbol{\mu}}_t
&= \frac{\sqrt{\alpha_t}(1 – \bar{\alpha}_{t–1})}{1 – \bar{\alpha}_t} \mathbf{x}_t + \frac{\sqrt{\bar{\alpha}_{t–1}}\beta_t}{1 – \bar{\alpha}_t} \frac{1}{\sqrt{\bar{\alpha}_t}}(\mathbf{x}_t – \sqrt{1 – \bar{\alpha}_t}\boldsymbol{\epsilon}_t) \\
&= \frac{1}{\sqrt{\alpha_t}} \Big( \mathbf{x}_t – \frac{1 – \alpha_t}{\sqrt{1 – \bar{\alpha}_t}} \boldsymbol{\epsilon}_t \Big)
\end{aligned}
\end{equation} 
where $\boldsymbol{\epsilon}_t$ denotes the noise involved in the forward diffusion process.

\begin{algorithm}[tb]
	\caption{Sampling of TDiffDe}
	\label{al2}
	\begin{algorithmic}[1]
		\Require
		{The noised hyperspectral image $\tilde{\mathbf{x}}$, the denoising step $T_{cut}<T$, the hyperparameters $\{\beta_1,\dots,\beta_T\}$} in diffusion training
		\Ensure {The denoised hyperspectral image $\mathbf{x}_0$}
		\State{Calculate $\{\alpha_1,\dots,\alpha_T\}$ following $\alpha_t = 1 – \beta_t$}
		\State{$t=T_{cut};~\mathbf{x}_t=\tilde{\mathbf{x}}$}
		\For {$t>0$}
		\If{$t>1$}
		\State{Randomly sample a Gaussian noise $\mathbf{z}_t \sim \mathcal{N}(\mathbf{0}, \mathbf{I})$}
		\Else
		\State{$\mathbf{z}_t=\mathbf{0}$}
		\EndIf
		\State{\textbf{end}}
		\State{Calculate $\bar{\alpha}_t = \prod_{i=1}^t \alpha_i$}
		\State{Calculate $\sigma_t= \sqrt{\frac{1 – \bar{\alpha}_{t–1}}{1 – \bar{\alpha}_t} \cdot \beta_t}$}
		\State{Feed $\mathbf{x}_{t},~t$ into $\boldsymbol{\epsilon}_\theta$ and obtain the estimated noise $\hat{\boldsymbol{\epsilon}}_t$}
		\State{Update $\mathbf{x}_{t–1}=\frac{1}{\sqrt{\alpha_t}} \Big( \mathbf{x}_t – \frac{1 – \alpha_t}{\sqrt{1 – \bar{\alpha}_t}} \hat{\boldsymbol{\epsilon}}_t \Big)+\sigma_t\mathbf{z}_t$, $t=t–1$}
		\EndFor
		\State{\textbf{end}}
	\end{algorithmic}
\end{algorithm}
\begin{figure*}[t]
	\vspace{-20pt}
	\addtolength{\leftskip}{-10pt}
	\centering
	\begin{minipage}[t]{0.5\linewidth}
		\subfloat[Noi]{\includegraphics[width=.315\linewidth]{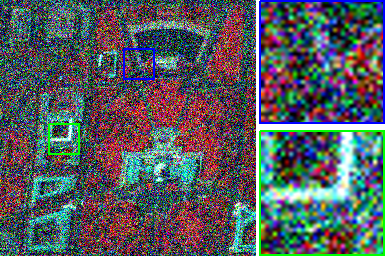}}\thinspace \thinspace
		\subfloat[LRMR]{\includegraphics[width=0.315\linewidth]{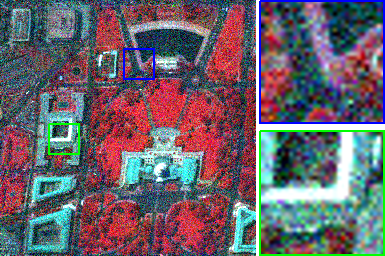}}\thinspace \thinspace
		\subfloat[LRTDTV]{\includegraphics[width=0.315\linewidth]{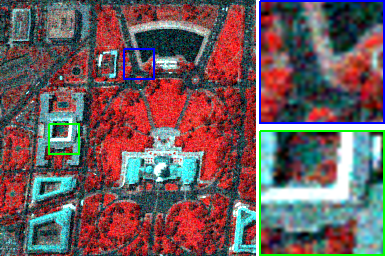}}
		\vspace{-10pt}
		\subfloat[CTV]{\includegraphics[width=0.315\linewidth]{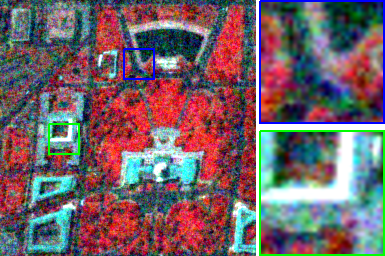}}\thinspace \thinspace
		\subfloat[RCTV]{\includegraphics[width=0.315\linewidth]{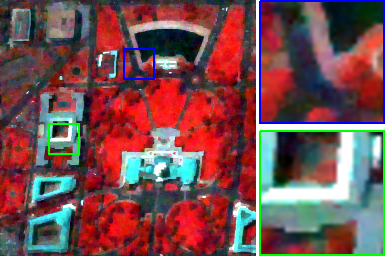}}\thinspace \thinspace
		\subfloat[NSSNN]{\includegraphics[width=0.315\linewidth]{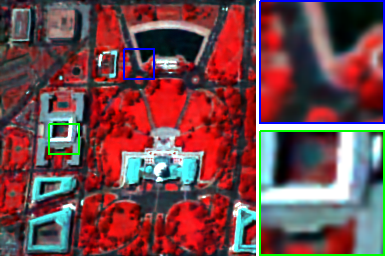}}
		\vspace{-10pt}
		\subfloat[QRNN3D]{\includegraphics[width=0.315\linewidth]{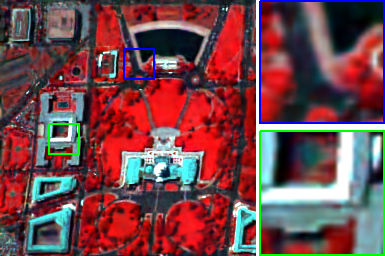}}\thinspace \thinspace
		\subfloat[\textbf{TDiffDe}]{\includegraphics[width=0.315\linewidth]{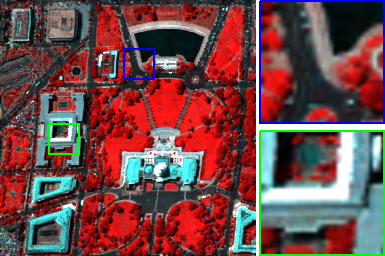}}\thinspace \thinspace
		\subfloat[GT]{\includegraphics[width=.315\linewidth]{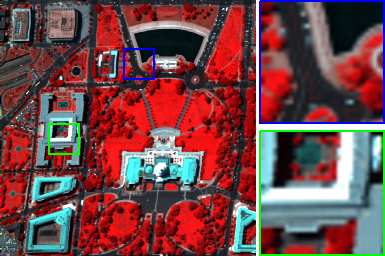}}\\
	\end{minipage}\thinspace
	\begin{minipage}[t]{0.5\linewidth}
		\subfloat[Noi]{\includegraphics[width=.315\linewidth]{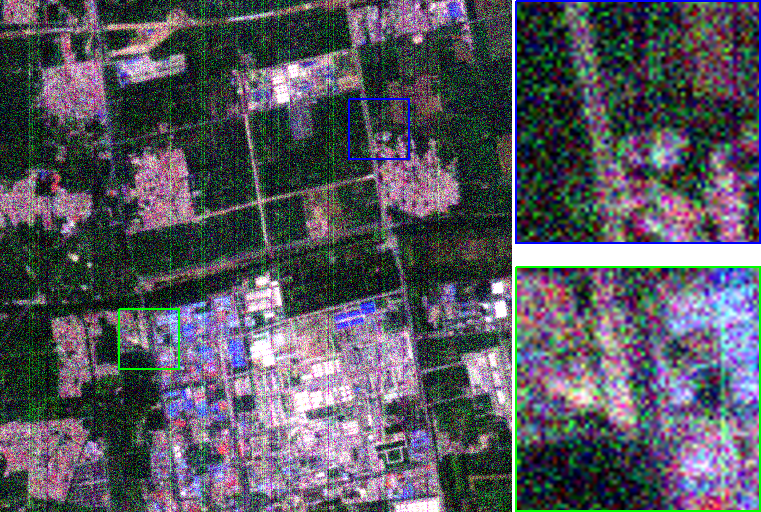}}\thinspace \thinspace
		\subfloat[LRMR]{\includegraphics[width=0.315\linewidth]{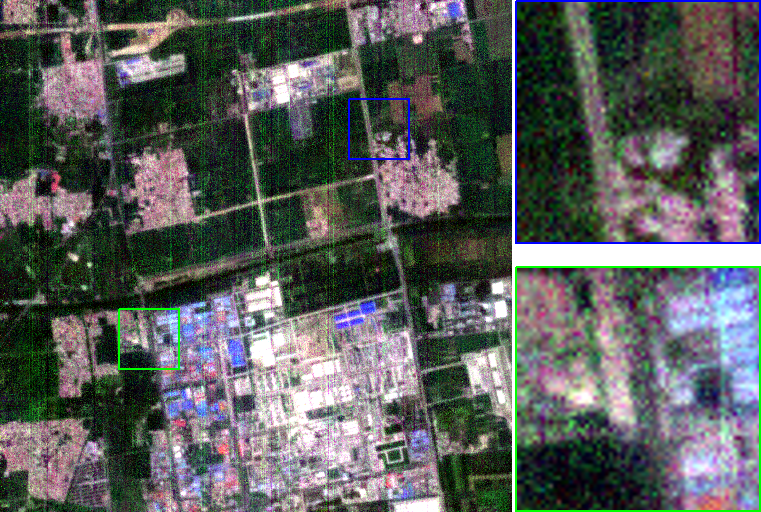}}\thinspace \thinspace
		\subfloat[LRTDTV]{\includegraphics[width=0.315\linewidth]{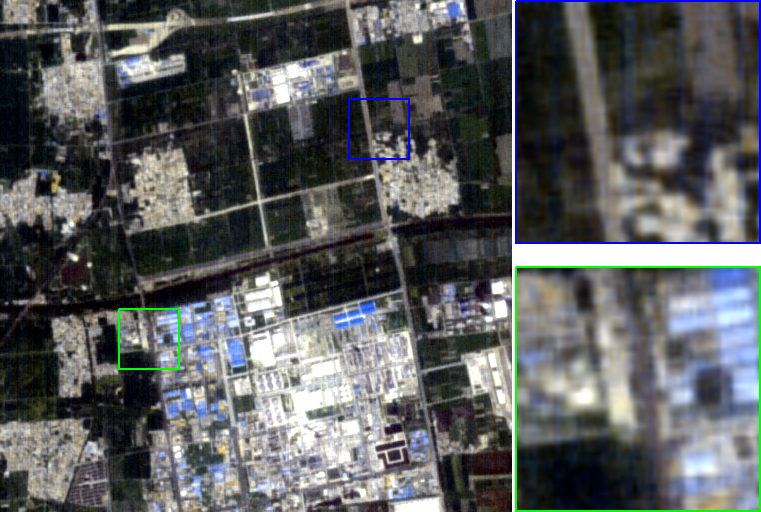}}
		\vspace{-10pt}
		\subfloat[CTV]{\includegraphics[width=0.315\linewidth]{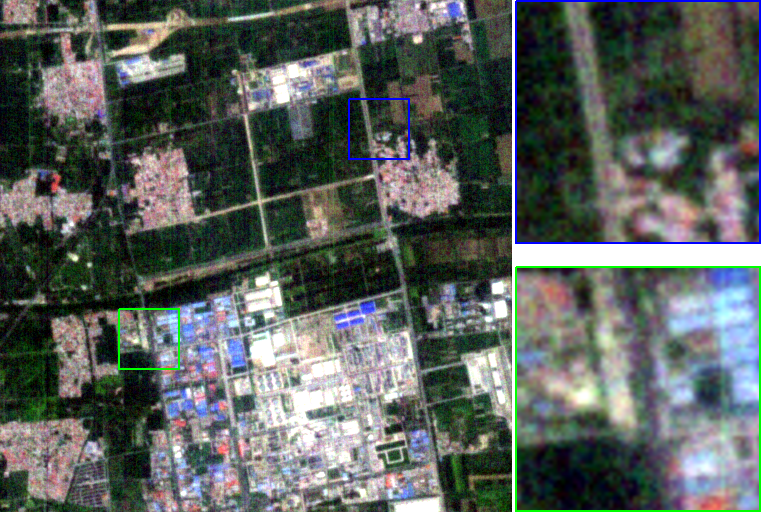}}\thinspace \thinspace
		\subfloat[RCTV]{\includegraphics[width=0.315\linewidth]{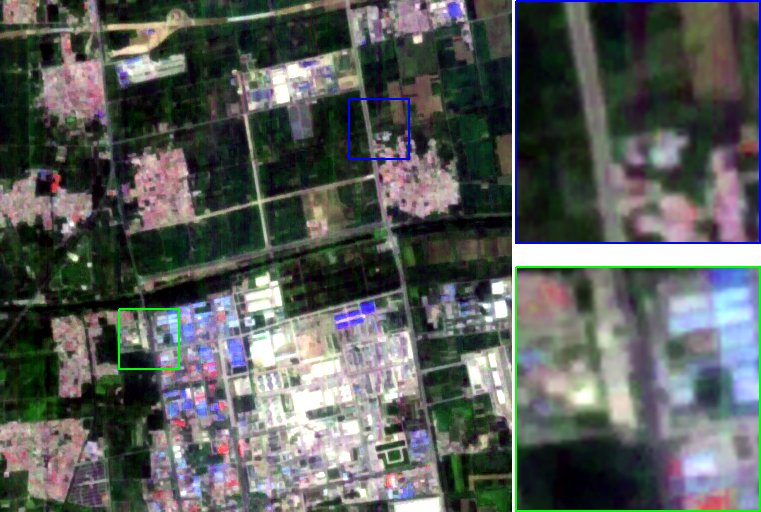}}\thinspace \thinspace
		\subfloat[NSSNN]{\includegraphics[width=0.315\linewidth]{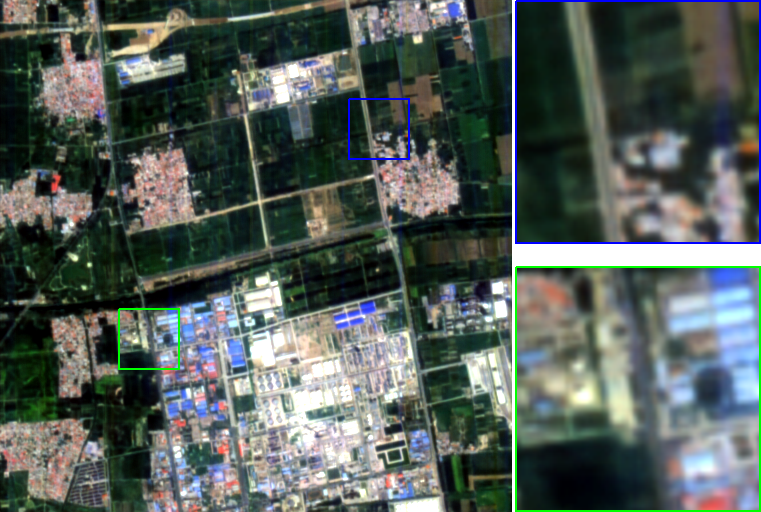}}
		\vspace{-10pt}
		\subfloat[QRNN3D]{\includegraphics[width=0.315\linewidth]{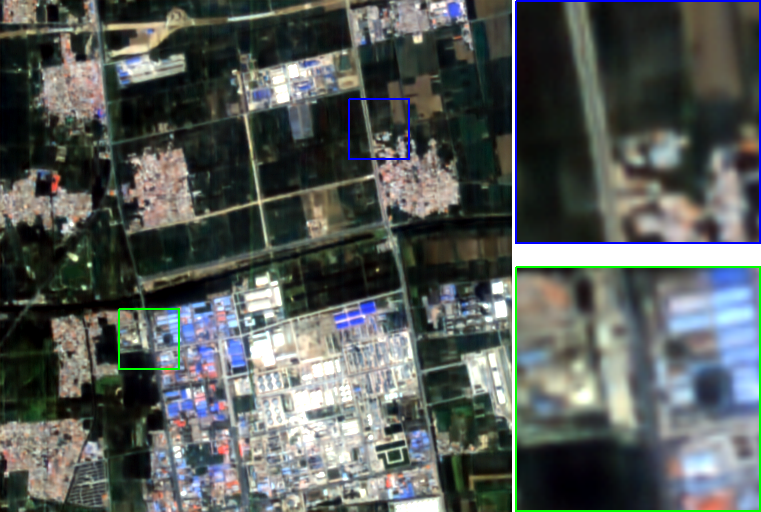}}\thinspace \thinspace
		\subfloat[\textbf{TDiffDe}]{\includegraphics[width=0.315\linewidth]{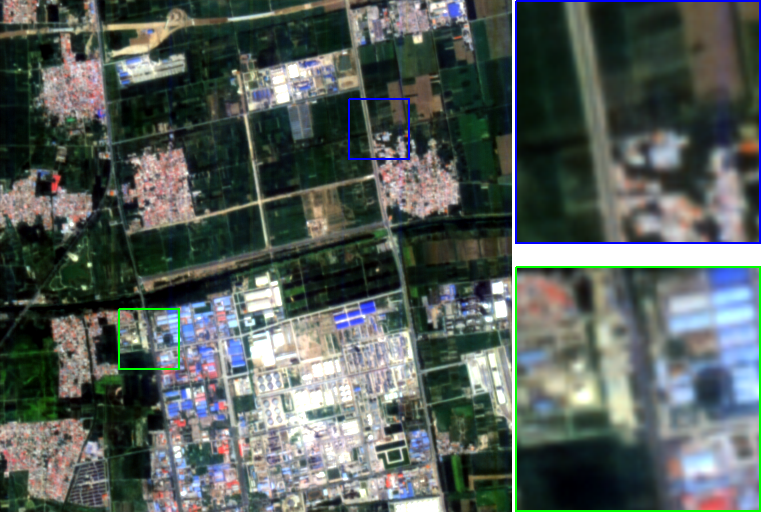}}\thinspace \thinspace
		\subfloat[GT]{\includegraphics[width=.315\linewidth]{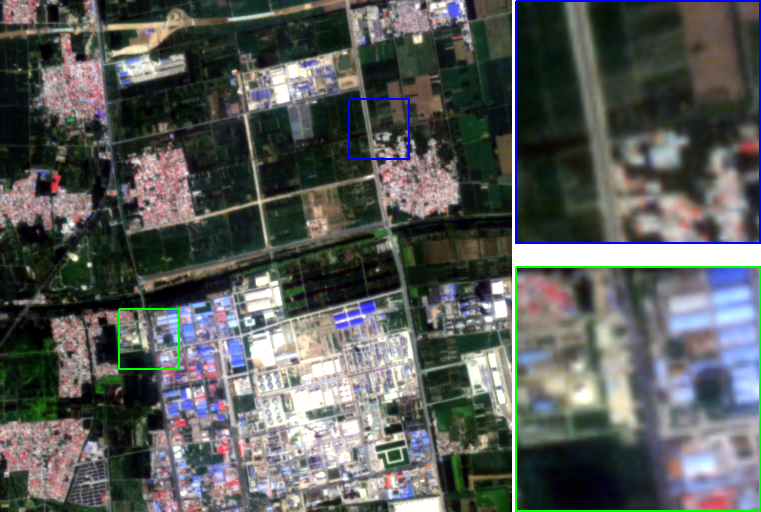}}
	\end{minipage}
	\caption{The left shows results on DC Mall dataset with $\sigma_n=50$. The right shows results on OHS dataset with hybrid noise.}\label{simu_results}\vspace{-10pt}
\end{figure*}

In hyperspectral image denoising, directly running reverse process would destroy the valid information in the noised hyperspectral image. In this paper, the reverse process in the diffusion model is truncated at $T_{cut}$ step as shown in Algorithm \ref{al2}. Furthermore, with a proper $T_{cut}$, the trained CNN–based noise predictor $\boldsymbol{\epsilon}_\theta$ can estimate the accurate noise level and take the noise out of images gradually.
\section{Experimental results}
To verify the performance of the proposed TDiffDe, we carried out both simulated and real experiments on two hyperspectral datasets. 
\subsection{Experimental setting}
\begin{table}[t]
\caption{Quantitative assessment of simulated experiments.}\label{tbl1}
	\centering
	\fontsize{8}{9}\selectfont
	\begin{tabular}{crrrrr}
		\toprule
		\begin{tabular}[c]{@{}l@{}}Noise\\Level\end{tabular} & Methods  & CC     & mPSNR   & mSSIM  & SAM      \\ \toprule
		\multirow{6}{*}{25}      & NOISE   & 0.6632          & 20.173           & 0.3575          & 29.0137         \\
		& LRMR    & 0.9219          & 29.4298          & 0.774           & 10.3343         \\
		& LRTDTV  & 0.9696          & 33.9944          & 0.8971          & 5.2182          \\
		& CTV & 0.9618	& 32.6018 &	0.8833 &	6.1623 \\ 
		& RCTV &0.9691 &	33.3856 	&0.9038 	&5.3734 	\\
		& QRNN3D  & 0.9781          & {\ul 34.8790}           & 0.9367          & {\ul 3.9741}          \\
		& NSSNN   & {\ul 0.9835}    & 34.5007    & {\ul 0.9387}    & 4.5704    \\
		& TDiffDe & \textbf{0.9862} & \textbf{36.7465} & \textbf{0.9569} & \textbf{2.3483} \\ \midrule
		\multirow{6}{*}{50}     & NOISE   & 0.4251          & 14.1528          & 0.1473          & 46.1491         \\
		& LRMR    & 0.7843          & 23.5602          & 0.5195          & 19.5719         \\
		& LRTDTV  & 0.9086          & 28.6815          & 0.7358          & 9.5495          \\
		& CTV &0.8920 	&27.9272 	&0.7180 	&10.5527 \\
		& RCTV& 0.9182 	&29.2844 &	0.7840 &	8.6773 \\
		& QRNN3D  & 0.9604          & 30.867           & {\ul 0.8755}    & 6.6624          \\
		& NSSNN   & {\ul 0.961}     & {\ul 31.0242}    & 0.8699          & {\ul 5.4793}    \\
		& TDiffDe & \textbf{0.9702} & \textbf{33.0331} & \textbf{0.9096} & \textbf{3.4153} \\ \midrule
		\multirow{6}{*}{75}     & NOISE   & 0.3039          & 10.6265          & 0.0754          & 56.4469         \\
		& LRMR    & 0.6552          & 20.0896          & 0.3551          & 28.2168         \\
		& LRTDTV  & 0.8475          & 25.7932          & 0.615           & 13.898          \\
		& CTV & 0.8097 	& 24.9733 	& 0.5658 	& 14.8320 \\
		& RCTV &0.8625 	&26.9817 	&0.6774 	&11.3449 \\
		& QRNN3D  & 0.9165          & 27.8398          & 0.7529          & 7.5503          \\
		& NSSNN   & \textbf{0.9311} & {\ul 28.7305}    & {\ul 0.7845}    & {\ul 6.1852}    \\
		& TDiffDe & {\ul 0.9246}    & \textbf{29.0817} & \textbf{0.7901} & \textbf{5.5656} \\ \midrule
		\multirow{6}{*}{Hybrid} & NOISE   & 0.8554          & 25.6577          & 0.5168          & 12.7224         \\
		& LRMR    & 0.9479          & 30.7662          & 0.7518          & 6.5098          \\
		& LRTDTV  & 0.9810          & 25.8345          & 0.8122          & 6.9934          \\
		& CTV & 0.9860 	& 36.0807 	& 0.9015 	& {\ul 2.8008} \\
		& RCTV &0.9887 	&{\ul 36.9123} 	&0.9399 	&\textbf{2.4466} \\
		& QRNN3D  & 0.9903          & 34.6180          & 0.9528          & 4.1384          \\
		& NSSNN   & {\ul 0.9906 }         & 36.7622         & {\ul 0.9608}          & 3.9719         \\
		& TDiffDe &  \textbf{0.9910} 	&	\textbf{36.9650} 	&	\textbf{0.9788} 	&	3.8160 
		\\ \bottomrule
	\end{tabular}
	\vspace{-10pt}
\end{table}

In this work, we used two datasets in the experiments. The first one is Washington DC Mall (DC Mall) data set \footnote{\url{https://engineering.purdue.edu/biehl/MultiSpec/hyperspectral.html}}. It is a widely used hyperspectral image data with a size of $1280 \times 307 \times 210$, covering the spectral wavelength from 400 nm to 2500 nm. The second one is a satellite image data set consisting of images from Orbita hyperspectral satellites (OHS) with 10–m spatial resolution and 32 bands\footnote{\url{https://zenodo.org/record/5642597}}. 

Five classical HS image denoising algorithms are compared in our paper, including LRMR \cite{lrmr}, LRTDTV \cite{lrtdtv}, CTV, RCTV, QRNN3D \cite{qrnn3d} and NSSNN \cite{nssnn}. Besides, correlation coefficient (CC), mean peak signalto–noise ratio (mPSNR), mean structural similarity (mSSIM), and spectral angle mapper (SAM), are utilized to evaluate the performance of all comparison methods quantitatively.

\begin{figure}[t]
	\centering
	\includegraphics[height=120pt]{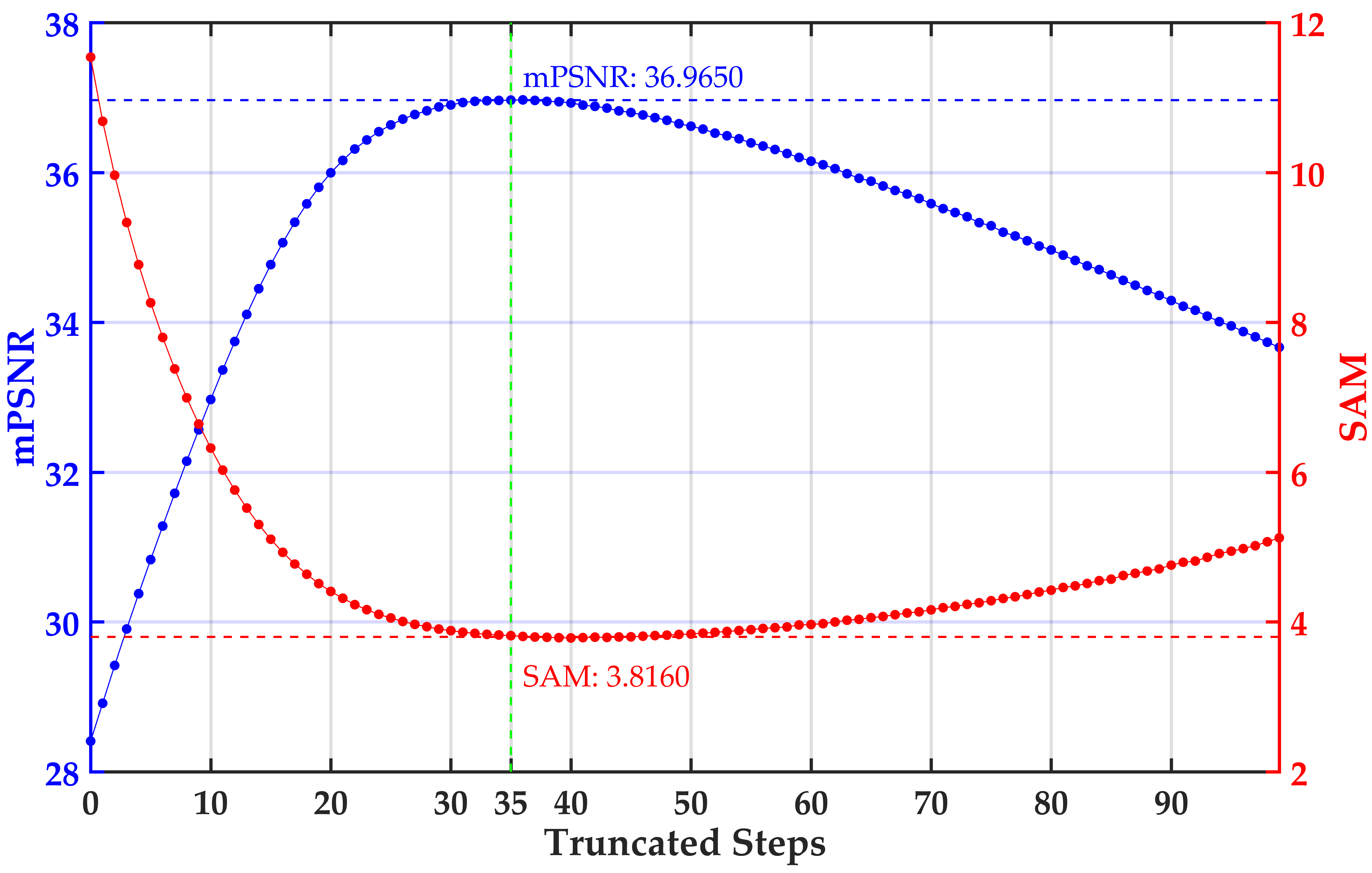}
	\vspace{-5pt}
	\caption{Sensitivity analysis of the truncation step $T_{cut}$.}\label{sensitivity}
	\vspace{-15pt}
\end{figure}

\begin{figure*}[t]
	\centering
	\subfloat[Noisy]{
		\begin{minipage}{75pt}
			\centering
			\includegraphics[width=1.0\linewidth]{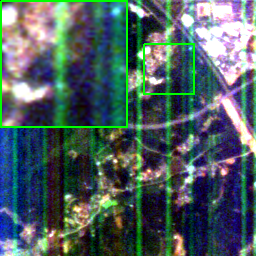}
		\end{minipage}
	}\quad
	\subfloat[LRMR]{
		\begin{minipage}{75pt}
			\centering
			\includegraphics[width=1.0\linewidth]{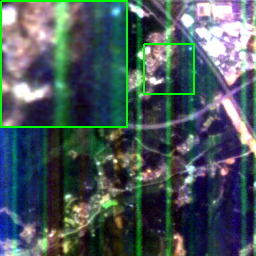}
		\end{minipage}
	}\quad
	\subfloat[LRTDTV]{
		\begin{minipage}{75pt}
			\centering
			\includegraphics[width=1.0\linewidth]{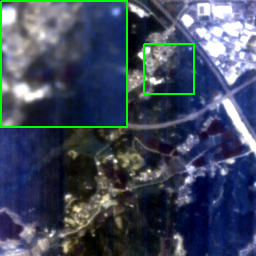}
		\end{minipage}
	}\quad
	\subfloat[CTV]{
		\begin{minipage}{75pt}
			\centering
			\includegraphics[width=1.0\linewidth]{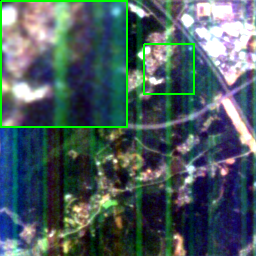}
		\end{minipage}
	}\\
	\subfloat[RCTV]{
		\begin{minipage}{75pt}
			\centering
			\includegraphics[width=1.0\linewidth]{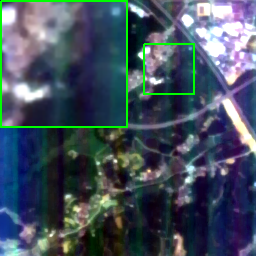}
		\end{minipage}
	}\quad
	\subfloat[QRNN3D]{
		\begin{minipage}{75pt}
			\centering
			\includegraphics[width=1.0\linewidth]{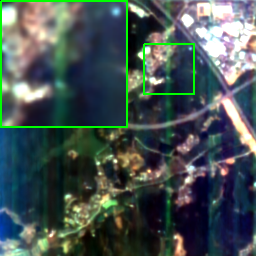}
		\end{minipage}
	}\quad
	\subfloat[NSSNN]{
		\begin{minipage}{75pt}
			\centering
			\includegraphics[width=1.0\linewidth]{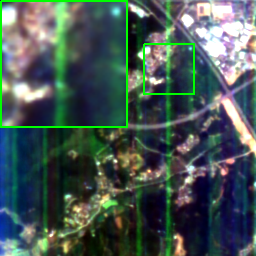}
		\end{minipage}
	}\quad
	\subfloat[\textbf{TDiffDe}]{
		\begin{minipage}{75pt}
			\centering
			\includegraphics[width=1.0\linewidth]{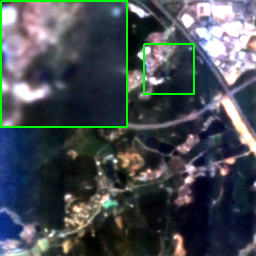}
		\end{minipage}
	}
	\caption{Results for the real noisy hyperspectral images.}\label{real_results}
\end{figure*}
In TDiffDe, we set the step $T=1000$, the hyperparameter $\beta_T=0.02,~\beta_1=0.0001$, and the truncated step $T_{cut}=35$. The adaptive moment estimation (Adam) with 0.0001 learning rate is employed to train TDiffDe. For CNN–based algorithms, models are all coded in the Pytorch framework and trained on a server equipped with a Nvidia RTX A5000 GPU and a 64 GB RAM. Moreover, the hyperparameter setting of the comparison models are all followed their paper. 
\subsection{Sensitivity analysis for $T_{cut}$}
Fig.~\ref{sensitivity} shows the change of PSNR performance as $T_{cut}$ setting from 0 to 100. mPSNR increases until $T_{cut}=35$, and the mPSNR is up to the maximum 36.9650. Considering computational cost, we choose $T_{cut}=35$ for our TDiffDe.

\subsection{Simulated experiments}
In this work, we carried out simulated experiments on both DC Mall and OHS data. The additive white Gaussian noise (AWGN) in different noise levels are added into DC Mall data over all bands, where $\sigma_n=\left[25,~50,~75\right]$. Moreover, hybrid noise are added into OHS data randomly,  consisting of Gaussian noise, impulse noise, and stripes.

Table~\ref{tbl1} reports the quantitative results with various noise. Various noise are used to train other deep learning–based methods on DC Mall and OHS datasets respectively, while only Gaussian noise is used in the training diffusion process of TDiffDe. Obviously, the proposed TDiffDe can achieve the best metric values when AWGN at different noise levels are added. With hybrid noise, TDiffDe can still perform good metric values. Fig.~\ref{simu_results} displays the visual results on DC Mall data with noise level $\sigma_n=50$ and OHS data with hybrid noise. The proposed TDiffDe can produce images representing high consistency with the ground truth.
\subsection{Real experiments}
In this part, we present the real experiments on the real OHS data with noise. The models used in simulated experiments are utilized directly. Fig.~\ref{real_results} shows the visual performance of different algorithms. LRMR can hardly remove the real noise. TDiffDe performs best, which effectively removes not only Gaussian noise but also strips. 
\section{Conclusion}
This study shows a new approach to achieve hyperspectral image denoising with a truncated diffusion model named TDiffDe. By adding Gaussian noise into images step–by–step in training diffusion process, TDiffDe can perform well on AWGN removal during reverse denoising. Results on DC Mall dataset have proved the validity of TDiffDe. Furthermore, as the results on OHS dataset representing, TDiffDe can also address hybrid noise, including stripes and impulse noise. Thus, as an image generation algorithm, diffusion–based models have enormous potential to remove hybrid noise with only Gaussian noise as the guide. Furthermore, it shows that the hybrid noise obeys the asymmetric Laplacian distribution. Introducing this distribution into DDPM is our future work.

{\small
\bibliographystyle{IEEEtran}
\bibliography{mybibfile}
}

\end{document}